\documentclass{article}

\PassOptionsToPackage{numbers, compress}{natbib}

\usepackage[final]{neurips_2024}
\usepackage{util}
\usepackage{multirow}
\usepackage{tabularx}

\usepackage[utf8]{inputenc} 
\usepackage[T1]{fontenc}    
\usepackage{hyperref}       
\usepackage{url}            
\usepackage{booktabs}       
\usepackage{amsfonts}       
\usepackage{nicefrac}       
\usepackage{microtype}      
\usepackage{xcolor}         

\title{Better Sampling, towards Better End-to-end Small Object Detection}

\author{
Zile Huang \And
Chong Zhang \And
Mingyu Jin \And
Fangyu Wu \And
Chengzhi Liu \And
Xiaobo Jin\footnote{} \\
Xi'an Jiaotong-Liverpool University \\
{Zile.Huang21, Chong.zhang19, Mingyu.jin19}@student.xjtlu.edu.cn, \\
{Fangyu.Wu02, Xiaobo.Jin}@xjtlu.edu.cn}

\begin{document}

\maketitle

\begin{abstract}
 While deep learning-based general object detection has made significant strides in recent years, the effectiveness and efficiency of small object detection remain unsatisfactory. This is primarily attributed not only to the limited characteristics of such small targets but also to the high density and mutual overlap among these targets. The existing transformer-based small object detectors do not leverage the gap between accuracy and inference speed.
 To address challenges, we propose methods enhancing sampling within an end-to-end framework. Sample Points Refinement (SPR) constrains localization and attention, preserving meaningful interactions in the region of interest and filtering out misleading information. Scale-aligned Target (ST) integrates scale information into target confidence, improving classification for small object detection. A task-decoupled Sample Reweighting (SR) mechanism guides attention toward challenging positive examples, utilizing a weight generator module to assess the difficulty and adjust classification loss based on decoder layer outcomes.
 Comprehensive experiments across various benchmarks reveal that our proposed detector excels in detecting small objects. Our model demonstrates a significant enhancement, achieving a 2.9\% increase in average precision (AP) over the state-of-the-art (SOTA) on the VisDrone dataset and a 1.7\% improvement on the SODA-D dataset.
\end{abstract}

\section{Introduction}

Since the emergence of deep convolutional neural networks (CNN), the performance of object detection methods has improved rapidly. There are two main approaches: two-stage proposal-based models with accuracy advantages \cite{dai2016-rfcn}  and single-stage proposal-free models with speed advantages \cite{cheng2017-dssd}. However, despite recent tremendous advances in object detection, detecting objects under certain conditions is still difficult, such as being small, occluded, or truncated. 


\begin{figure}[htp] 
    \centering 
    \includegraphics[width=1.0\linewidth]{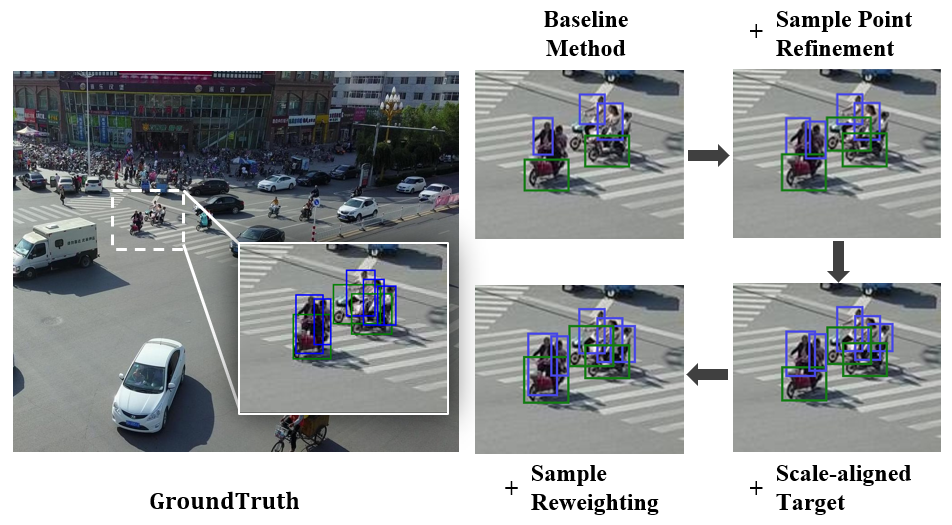} 
    \caption{Visualization of our methods within challenging scenes featuring crowds and overlapping objects. Our visual illustration emphasizes the effectiveness of scale-aligned target, sample reweighting, and sample point refinement within intricate scenes, allowing for a comparative analysis with a standard or baseline method.}
    \label{size imbalance within and between classes visualization}
\end{figure}


Small target detection problems are mainly divided into four categories: multi-scale representation, contextual information, super-resolution and region proposal. Due to the tiny size and low resolution of small objects, location details are gradually lost in high-level feature maps. Multi-scale representation \cite{cui2020mdssd} combines the location information of low-level features and the semantic information of high-level features.  Contextual information, which exploits the relationship between an object in an image and its surrounding environment, is another novel way to improve small object detection accuracy. It is very necessary to extract more additional contextual information \cite{liu2021modular} as a supplement to the original ROI (Region of Interest) features because there are too few ROI features extracted from small objects.  As mentioned above, fine details are crucial for object instance localization, which means that more details for small objects should be obtained.  Super-resolution technology \cite{cao2017learning} attempts to restore or reconstruct the original low-resolution image to a higher resolution, which means more details of small objects can be obtained. Region-proposal \cite{chen2019ssdmsn} is a strategy aimed at designing more suitable anchor points for small objects. The anchors of the current leading detectors mainly focus on generic objects, and the size, shape and number of anchors used in general detectors do not match small objects well.


Through the Transformer architecture, DETR \cite{carion2020end} achieves competitive end-to-end detection performance in a one-to-one scheme, abandoning the manual components of NMS and eliminating the need for post-processing. However, the limited number and low quality of positive samples may lead to poor performance on objects of extremely limited size. Deformable DETR \cite{zhu2020deformable} uses deformable sampling points to enhance its feature decoding capabilities, but its performance in densely populated and overlapping scenes is unsatisfactory, and the ongoing challenge of effectively distinguishing positive and negative samples remains solve.


To address the inherent challenges associated with small object detection and the limitations observed in current end-to-end object detectors, we introduce three different strategies aimed at mitigating these issues from different perspectives. In this work, we focus on improving the DETR method \cite{carion2020end} for detecting small target objects. First, to mitigate potential clutter caused by background noise and surrounding elements, our approach follows an iterative refinement paradigm similar to \cite{zhu2020deformable}, which focuses on refining the attention mechanism by merging updated box predictions from the previous layer, and effectively reduce confusion with background elements and noise. Second, sample point refinement (SPR) constrains the distribution and attention of sample points by setting learning goals for the deformable attention module in the decoder, instead of directly discarding reference points located outside the bounding box. This intentional inhibition can serve as a strategic stimulus for the model, resulting in a more object-centered deformable attentional sampling capability without losing surrounding information. Finally, a decoupled sample reweighting (SR) mechanism integrates a scaling factor into the target confidence, making the model inherently scale-aligned. The decoupled reweighting technique helps the model refocus its focus to learning from challenging positive examples at the feature level instead of the object level.


In summary, the main contributions of this article are as follows:

\im{
\item We proposed Sample Points Refinement (SPR), which imposes constraints on the localization and attention of sample points within deformable attention. This ensures the preservation of meaningful interactions within the region of interest while filtering out potentially misleading information.

\item We incorporate scale information from samples into the target confidence, namely Scale-aligned Target (ST), thereby establishing a more appropriate learning objective for classification. This adjustment proves highly advantageous, particularly in the realm of small object detection.

\item We present a task-decoupled Sample Reweighting (SR) mechanism aimed at guiding the model to redirect its attention toward learning from challenging positive examples. Our approach includes a weight generator module that gauges difficulty from decoder layer outcomes, facilitating the adjustment of the classification loss accordingly.

\item We conduct comprehensive experiments and ablations on VisDrone \cite{du2019visdrone} and SODA-D \cite{Cheng2022TowardsLS} to verify the effectiveness of the proposed method and the importance of correct sampling in small object detection.
}

\section{Related work}

\paragraph{Multi-scale Feature Learning} The basic idea of the multi-scale feature learning method is to learn targets at different scales separately, mainly to solve the problem that small targets themselves have few discriminative features, including based on feature pyramids and based on receptive fields. The main idea of the feature pyramid-based method \cite{guo2020augfpn} is to integrate low-level spatial information and high-level semantic information to enhance target features. Receptive field-based methods, such as the Trident network \cite{li2019scaleaware}, use dilated convolutions with different expansion rates to form three branches with different receptive fields, responsible for detecting objects of different scales.

\paragraph{GAN-based Methods} Different from multi-scale feature learning methods, GAN methods solve the problem of small targets with few discriminative features by generating high-resolution images or high-resolution features. SOD-MTGAN \cite{bai2018sod-mtgan} uses a trained detector (such as Faster R-CNN) to obtain a subgraph containing objects, and then uses a generator to generate corresponding high-definition images, where the detector also acts as a discriminator. After generating high-resolution features, GaN network \cite{Noh2019BetterTF} reduce the input image by 2 times, then extracts the features as low-resolution features, and the features extracted from the original image as the corresponding high-resolution features. The generator generates ``fake" high-resolution features based on low-resolution features, and the discriminator is responsible for distinguishing ``fake" high-resolution features from ``real" high-resolution features.

\paragraph{Context-based Methods  }  Context-based methods assist the detection of small objects based on the environmental information where the small objects are located or their relationship with other easily detected objects. Inside-Outside Net \cite{bell2016inside-outside} extracts the global context information of each object in the four directions of up, down, left, and right through RNN. RelationNetwork \cite{han2018relation} uses the relationship between objects to assist in the detection of small objects. The relationship between two objects is implicitly modeled by a Transformer, and this relationship is used to enhance the characteristics of each object.

\paragraph{Loss Reweighting Methods} This type of method makes the network pay more attention to the training of small objects by increasing the weight in the small object loss. The feedback-driven loss method \cite{liu2021feedback-driven} increases the weight of small object loss when the proportion of small object loss is low, allowing the network to treat objects of different scales more equally.

\paragraph{Special Design for Small Object} This type of method designs target detection algorithms based on the characteristics of small targets. S3FD \cite{zhang2017s3fd} directly reduces the IoU threshold of small object positive samples, thereby increasing the number of anchors for small object matching. If there are still few anchors matching small objects, the top $N$ anchors are selected from all anchors that meet the threshold as matching anchors.  Xu et al. \cite{xu2021dot} proposed a new metric to alleviate the situation where a slight shift in the prediction box of a small object causes a huge change in the IoU index.  Wang et al. \cite{wang2022normalized} introduced the normalized Wasserstein distance to optimize the position metric for small object detectors. However, this method is not sensitive to box offset, and a small amount of offset will not cause a sharp drop in the indicator.

\section{Background on Deformable DETR-based Detection}

\ssn{Deformable DETR-based Models}

Following the Deformable DETR-based models \cite{zhu2020deformable}, our model consists of three main components: a CNN-backbone, an encoder-decoder transformer, and a prediction head.

The backbone network flattens the raw features of the input image into a sequence of tokens $X = \{x_1,x_2,\cdots,x_l\}$. The transformer then extracts information from $X$ using a set of learnable queries $Z = \{z_1,z_2,\cdots,z_n\}$. 

We can compute the attention representation for the i$^{th}$ query point
\eqna{
&& \tm{Attn}(z_i, p_i,X) \nn \\
& = & \sum_{m} W_m [\sum_{l,k} A_{milk} W'_m x_l (\phi_l (p_i) + \Delta p_{milk})],\nn \\
}
where the set of subscripts $\{m,i,l,k\}$ represents the $k$-th sampling point of the $l$-th sampling layer of the $m$-th attention head corresponding to the $i$-th query and the reference point of the $i$-th query is $p_i$. The matrices $W$ and $W'$ correspond to two linear transformations respectively. Here, $\Delta p$ and $A$ denote the tensor matrices of sampling offsets and attention weights, respectively, where the weight matrix $A$ satisfies the constraints $\sum_{l,k}A_{milk} = 1$. It is worth noting that $p_i$ is a normalized two-dimensional coordinate located at $[0,1]^2$, and the function $\phi$ will convert it to the coordinates in the original image.

Subsequently, for each query $z_i$, the decoder will output two heads: the head $H^{\tm{reg}}(z_i)$ predicts the location of the object box in the image, and another head $H^{\tm{cls}}(z_i)$ predicts the category of the object
\begin{equation}
s_i = H^{\tm{cls}}(z_i), \quad b_i = H^{\tm{reg}}(z_i)
\label{head}
\end{equation}
where $s_i$ represents the object category corresponding to the $i$-th query, and $b_i$ is the box predicted by multiple decoders, including the upper left corner position and the length and width of the box.

\ssn{Varifocal Loss}

We adopt Varifocal Loss (VFL) \cite{zhang2021varifocalnet} as our baseline method for classification loss. Varifocal Loss is an extension of Focal Loss \cite{lin2017focal}, incorporating a trade-off parameter $\alpha$ that adjusts the loss based on the difficulty of each sample:
\begin{equation}
\hat{\mathcal{L}}_{\tm{cls}} = - \sum_{i = 1}^{N^{+}} q_i \tm{BCE}(p_i,q_i) - \alpha \sum_{i = 1}^{N^{-}} p_i^{\gamma} \tm{BCE}(p_i, 0),
\label{eqn:cls-loss}
\end{equation}
where $p_i$ is the predicted IoU-aware classification score (IACS) and $q_i$ is the target score. For a foreground point, $q_i$ is set to the IoU between the generated bounding box and its ground-truth value, otherwise it is $0$. For background points, the target $q_i$ is 0 for all classes. The variables $N^{+}$ and $N^{-}$ respectively represent the number of bounding boxes containing the object.

We perform bounding box regression using a combination of norm loss $\mc{L}_1$ and Intersection over Union (IoU) loss $\mc{L}^{\tm{IoU}}$, where $L1$ loss measures the absolute difference in width, height, center $x$, and center $y$ between the predicted bounding box and the ground truth, a measure of localization accuracy, while IoU computes the intersection area between the predicted and ground-truth bounding boxes divided by their union as a similarity measure and penalizes differences in spatial overlap
\begin{equation}
\hat{\mathcal{L}}^{\tm{reg}} = \sum_{i = 1}^{N^{+}} [\mathcal{L}_{1}(b_{i}, \hat{b}_{i}) +  \mathcal{L}^{\tm{IoU}}(b_{i},\hat{b}_{i})],
\label{eqn:reg-loss}
\end{equation}
where $b_i$ and $\hat{b}_i$ represent the predicted bounding boxes and the ground truth bounding box respectively. 
The total loss will be presented in the following section.

\sn{Our Method}

\begin{figure*}[tb] 
    \centering 
    \includegraphics[width=1\linewidth]{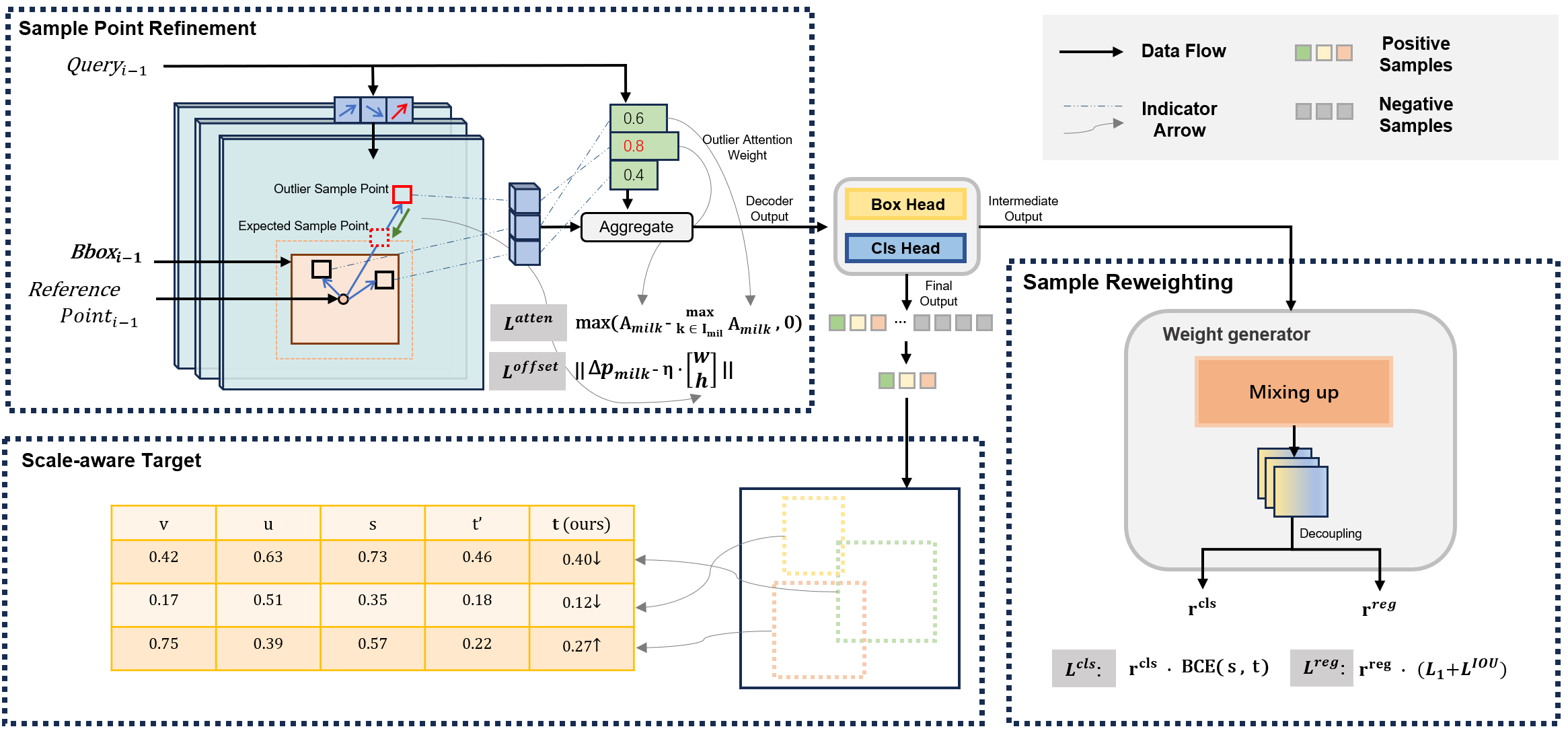} 
    \caption{Framework overall. We proposed a series of transferable methods to support end-to-end models to perform better in small object detection. The final loss could be represented as $\mc{L} = \mc{L}^{\tm{cls}} + \mc{L}^{\tm{reg}} + \mc{L}^{\tm{offset}} + \mc{L}^{\tm{atten}}$.}
    \label{Overall framework}
\end{figure*}
Our methodology involves multi-step training supervision aimed at refining sampling strategies, addressing challenges posed by box edge regression and ambiguous classification in small object detection, often caused by limited information availability. Notably, our approach maintains the original inference speed as we solely modify the training objective of the model.

\ssn{Refinement of Sample Points Outside Box}

We employ learnable offsets in deformable attention to make the convolution kernel more flexible to focus on objects with significant differences in size and shape. In Deformable DETR \cite{zhu2020deformable}, attention is focused on the x-coordinate and width of the left or right border of the object and the y-coordinate and height of the top or bottom border.  However, when the model is not constrained in learning deformation offsets, the perceptual field has a tendency to extend beyond the target, especially in the context of small objects. As in Fig \ref{Overall framework}, the red square within the deformable attention region signifies a sample point necessitating refinement due to its substantial distance from the object, resulting in diminished informativeness.

Suppose that when we use the $i$-th query to sample in the $l$-th layer of the $m$-th attention head, the set of sampling points contained within the bounding box is $\mc{I}_{ilm}$, and the set of sampling points located outside the bounding box is $\mc{O}_{ilm}$, then in order to make the points outside the bounding box as close as possible to the bounding box, we define the following loss for sampling points outside the bounding box
\eqn{}{
\mc{L}^{\tm{offset}} =  \sum_{i = 1}^{N^+} \left[ \sum_{m,l,k \in \mc{O}_{ilm} } \left \|\Delta p_{milk} - \eta \bmtx{w_i \\ h_i} \right \|_1 ^2 \right],
}
where $w_i$ and $h_i$ represent the width and height of the previous layer on the query $z_i$ in the decoder, and $\eta > 1$ is an expansion parameter indicating the expansion degree of the bounding box of the previous layer.  Note that since $\eta > 1$, the sample point is still outside the bounding box. The orange dotted line box in Fig \ref{Overall framework} functions as an expanded buffer, preserving both local and global attention cues.

At the same time, we hope that the weight $A$ of the sampling points outside the box should be smaller than the weight of all sampling points within the box, so we define the attention loss for sampling points outside the bounding box as follows
\eqna{
\mc{L}^{\tm{atten}} & = & \sum_{i = 1}^{N^+}  \sum_{m,l,k \in \mc{O}_{ilm} } \nn \\
&& \max ( A_{milk} - \max_{k \in \mc{I}_{mil} } A_{milk},0 ). \nn \\
}

\ssn{Estimation of Target Confidence}

\begin{figure}[htbp]
    \centering
    \begin{tikzpicture}
    \draw[red] (-1, -1) rectangle (1, 1);
    \fill[fill=gray!30] (0, 0) circle (1.0cm);
    \node[left] at (0, 0) { $c$ };
    \filldraw[red] (0, 0) circle (0.02cm);
    \draw[dotted, blue, thick] (0, 0) rectangle (1.5, -2.0);
    \draw[red] (2.0, -1) rectangle (4.0, 1);
    \fill[fill=gray!30] (3.0, 0) circle (1.0cm);
    \node[left] at (3.0, 0) { $c$ };
    \filldraw[red] (3.0, 0) circle (0.02cm);
    \draw[dotted, blue, thick] (2.7, 0.4) rectangle (4.0, -1.9);
\end{tikzpicture}

    \caption{The IoU values on the two target detection tasks are equal, but the right side has a larger area ratio (predicted box area and true box area).}
    \label{fig:enter-label}
\end{figure}
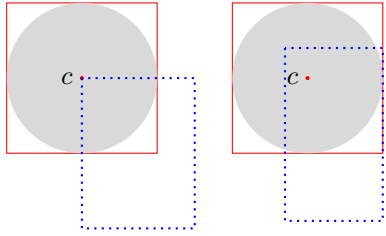

As shown in Eqn. (\ref{eqn:cls-loss}), although Varifocal loss increases the penalty for negative class samples, it does not consider the interaction between target detection and target classification. To this end, we introduce the results of small target detection and redefine the target confidence $s$.  For small target detection, detection performance is not only related to the intersection area between the predicted box and the grounding-truth box, but also related to the size of the predicted box. 

Assume that the annotations of the predicted box and the grounding-box box are $B = (x,y,w,h)$ and $\hat{B} = (\hat{x},\hat{y},\hat{w},\hat{h})$,  respectively, then we introduce two variables to measure the performance of the detection algorithm
\eqna{
u & = & \frac{\mc{A}(B \cap \hat{B}) }{ \mc{A} (B \cup \hat{B}) } = \tm{IoU}(B,\hat{B}),\\
\rho & = & \frac{\mc{A}(B) }{ \mc{A}(\hat{B})},
}
where $\mc{A}(\cdot)$ represents the area of the bounding box. If we define $x = \mc{A}(B)$, $y = \mc{A}(\hat{B})$ and $z = \mc{A}(B \cap \hat{B})$, then we have 
\eqna{
u & = & \frac{z}{ x + y - z}, \\
\rho & = & \frac{x}{y}.
}
Furthermore, we get the relationship between $u$ and $\rho$
\eqn{}{
\frac{z/y}{\rho + 1} = \frac{u}{1 + u}.
}
When IoU ($u$) is constant, $z/y \propto (\rho + 1)$. But obviously, we prefer larger $\rho$ or $z/y$, which indicates better detection results, since the area of the predicted box is closer to the area of the grounding-truth box.

Based on the above analysis, we define a function with respect to IoU $u$ and area ratio $\rho$ to re-represent the target confidence score
\eqna{
   r & = & \sqrt{\rho}, \label{eqn:sqrt-opt}\\
   v & = & e^{-\theta (r - 1)^2},\label{eqn:exp-opt}\\
   c & = & u^{\beta} \cdot v^{1 - \beta},\label{eqn:geom-opt}\\
   t & = & c\cdot s,
}
where $\beta$ and $\theta$ are the hyper-parameters. Note that our goal is that the area of the predicted box is as close as possible to the area of the true box, that is, $\rho \ra 1$. Therefore, we interpret the meaning of the above formula as follows: the square root operation in Eqn. (\ref{eqn:sqrt-opt}) is to make the values near $1$ more concentrated; the exponential operation in Eqn. (\ref{eqn:exp-opt}) is to convert the distance loss into a bounded interval $[0,1]$; Eqn. (\ref{eqn:geom-opt}) combines the indicators IoU ($u$) and $\rho$ through geometric mean as the proportional coefficient of $s$, where the geometric mean rather than the algorithmic mean is used to measure measurements on two different scales.

\ssn{Reweighting of Positive Samples}

\begin{figure}[tb]
    \centering
    \includegraphics[width=0.95\linewidth]{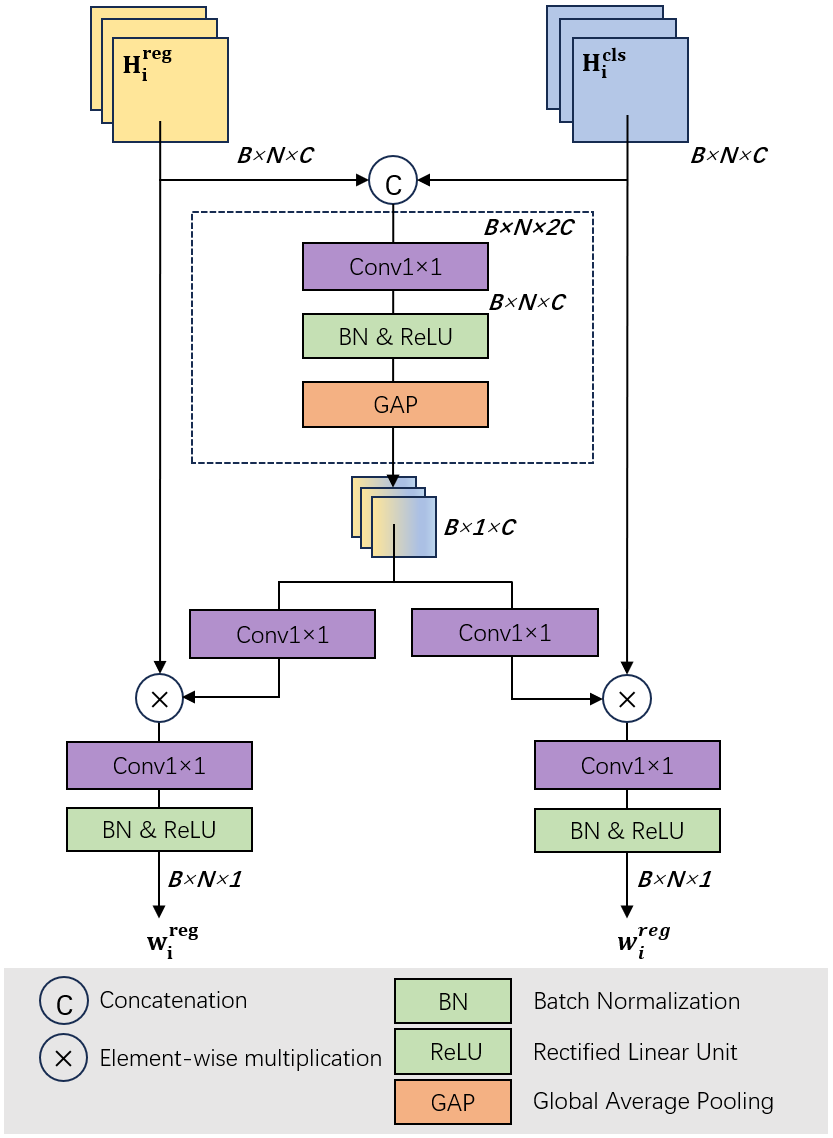}
    \caption{Illustration for the weight generator in the reweighting module. The kernel size of Conv$_{1}$, Conv$_{2}$, and Conv$_{3}$ are $C\times2C\times1\times1$, $C\times C\times1\times1$, and $1\times C\times1\times1$, respectively.}
    \label{fig:reweighting}
\end{figure}

Below we describe in detail our reweighting module for positive samples (boxes containing targets). 

First, we generate a shared attention feature between the classification and the regression tasks through the hidden layer representation of the two tasks: the inputs $H^{\tm{cls}}_i$ and $H^{\tm{reg}}_i$ respectively come from the hidden layer features decoded by the classification head and the regression head that decode the query $z_i$. Then, we directly concatenate the two inputs and go through the following operations, as shown in the dotted box in Fig \ref{fig:reweighting}. 
\eqn{}{
A_i = \sigma(\mc{B}(\tm{cov}(\tm{cat}(H^{\tm{reg}}_i,H^{\tm{cls}}_i)))),
}
where $\mc{B}$ represents ReLU and batch normalization operations and $\sigma$ is the sigmoid function.

Based on the shared attention weight $A_i$, We implement channel compression through convolution operation and BN/ReLU operation to obtain the weights $w^{\tm{cls}}_{i}$ and $w^{\tm{reg}}_i$ corresponding to query $z_i$ 
\eqna{
w^{\tm{cls}}_i & = & \sigma(\mc{B}(\tm{cov} (H^{\tm{cls}}_i \otimes \tm{cov} (A_i))), \\
w^{\tm{reg}}_i & = & \sigma(\mc{B}(\tm{cov} (H^{\tm{reg}}_i \otimes \tm{cov} (A_i))),
}
where $\otimes$ is the element-wise multiplication of matrices.

However, in the initial stage of training, when the model has not fully learned accurate feature representations, we discount the weights, where the weights are a decreasing function of the difference between target confidence and prediction confidence
\eqn{}{
r =  w^{1-|t - s|},
}
where $w$ can be $w^{\tm{cls}}_i$ or $w^{\tm{reg}}_i$ related to query $z_i$, $t$ and $s$ are respectively the target confidence and prediction confidence corresponding to $z_i$. Note that when our algorithm converges, then $t = s$ and $r = w$. Finally, we get the classification weight $r^{\tm{cls}}_i$ and regression weight $r^{\tm{reg}}_i$ of each bounding box containing the object.

Based on the above discussion,  the reweighted loss for both classification (\ref{eqn:cls-loss}) and regression (\ref{eqn:reg-loss}) is derived by substituting the suppression factor on positive samples with $r^{\tm{cls}}_i$, while retaining the suppression on negative samples.

\eqna{
\mathcal{L}^{\tm{cls}} & =  &  \sum_{i = 1}^{N^{+}} r^{\tm{cls}}_i \tm{BCE}(s_i,t_i) \nn \\
&& + \alpha \sum_{i = 1}^{N^{-}} p_i^{\gamma} \tm{BCE}(s_i, 0), \\
\mathcal{L}^{\tm{reg}} & = & \sum_{i = 1}^{N^{+}} r^{\tm{reg}}_i[\mathcal{L}_{1}(b_{i}, \hat{b}_{i}) +  \mathcal{L}^{\tm{IoU}}(b_{i},\hat{b}_{i})]. \nn \\
}

Finally, we train our object detection architecture using the following loss function
\eqn{}{
\mc{L} = \mc{L}^{\tm{cls}} + \mc{L}^{\tm{reg}} + \mc{L}^{\tm{offset}} + \mc{L}^{\tm{atten}}.
}

\begin{table*}[ht]
\centering
\caption{Comparison with state-of-the-art end-to-end detectors detection approaches on the Visdrone2019 test-set: The subscripts S, M, and L respectively indicate the size of the object as small, medium, and large.}
\resizebox{\textwidth}{!}{
\begin{tabularx}{1.2\textwidth}{lcccccccccc}
\toprule
Model & Publication & \#epochs & AP & $\mathrm{AP}_{50}$ & $\mathrm{AP}_{75}$ & $\mathrm{AP}_S$ & $\mathrm{AP}_M$ & $\mathrm{AP}_L$ & Params \\
\midrule
\textbf{Non-Deformable-DETR-based} &&&&&&&&&& \\
\midrule
YOLOX-R50 \cite{ge2021yolox}                  & CVPR'21 & 70 & 24.1 & 52.6 & 25.1 & 13.7 & 39.5 & 53.4 & $43 \mathrm{M}$ \\
Anchor-DETR-R50 \cite{wang2021anchor}         & AAAI'21 & 70 & 27.7 & 54.7 & 26.3 & 14.1 & 40.8 & 56.7 & $39 \mathrm{M}$ \\
Sparse RCNN-R50 \cite{sun2021sparse}          & CVPR‘21 & 12 & 27.2 & 56.9 & 25.7 & 17.4 & 46.2 & 57.1 & $40 \mathrm{M}$ \\
ViTDet-ViT-B \cite{li2022exploring}           & ECCV‘22 & 50 & 28.9 & 57.1 & 25.3 & 16.5 & 47.3 & 55.4 & $47 \mathrm{M}$ \\
\midrule
\textbf{Deformable-DETR-based} &&&&&&&&&& \\
\midrule
Deformable-DETR-R50 \cite{zhu2020deformable}  & ICLR'21 & 50 & 21.7 & 48.4 & 23.5 & 11.7 & 34.8 & 49.2 & $43 \mathrm{M}$ \\
Conditional-DETR-R50 \cite{meng2021-CondDETR} & ICCV'21 & 50 & 26.4 & 53.7 & 27.4 & 16.2 & 46.2 & 53.6 & $44 \mathrm{M}$ \\
DAB-DETR-R50 \cite{liu2022dabdetr}            & ICLR'22 & 50 & 27.8 & 56.8 & 26.9 & 16.4 & 45.5 & 54.8 & $47 \mathrm{M}$ \\
DINO-R50 \cite{zhang2022dino}                 & ICLR'23 & 36 & 26.8 & 58.2 & 28.9 & 17.5 & 47.3 & 58.3 & $47 \mathrm{M}$ \\
Co-DINO-R50 \cite{codetr2022}                 & ICCV‘23 & 36 & 29.4 & 58.7 & 29.5 & 18.9 & 46.8 & 61.3 & $55 \mathrm{M}$ \\
\midrule
Baseline-R50 \cite{Lv2023DETRsBY}             & -       & 72 & 28.9 & 57.3 & 31.1 & 19.3 & 46.0 & 57.3 & $42 \mathrm{M}$ \\
Ours-R50                                      & -       & 72 & $\textbf{32.3}$ & $\textbf{66.4}$ & $\textbf{36.7}$ & $\textbf{23.7}$ & $\textbf{50.0}$ & $\textbf{60.2}$ & $43 \mathrm{M}$\\
\bottomrule
\end{tabularx}
}
\label{visdrone}
\end{table*}

\begin{table*}[ht]
\centering
\caption{Comparison with state-of-the-art end-to-end detection approaches on the SODA-D test-set, where the subscripts ES, RS and GS represent extremely small, relatively small, generally small and normal respectively.}
\resizebox{\textwidth}{!}{
\begin{tabularx}{1.3\textwidth}{lcccccccccc}
\toprule
Model & Publication & \#epochs & AP & $\mathrm{AP}_{50}$ & $\mathrm{AP}_{75}$ & $\mathrm{AP}_{ES}$ & $\mathrm{AP}_{RS}$ & $\mathrm{AP}_{GS}$ & $\mathrm{AP}_{N}$ & Params \\
\midrule
\textbf{Non-Deformable-DETR-based} &&&&&&&&&& \\
\midrule
YOLOX-R50\cite{ge2021yolox}                  & CVPR'21 & 70 & 26.7 & 53.4 & 23.0 & 13.6 & 25.1 & 30.9 & 30.4 & $43 \mathrm{M}$ \\
Anchor-DETR-R50\cite{wang2021anchor}         & AAAI'21 & 70 & 27.3 & 55.9 & 22.7 & 12.2 & 24.2 & 32.8 & 41.7 & $39 \mathrm{M}$ \\
Sparse RCNN-R50\cite{sun2021sparse}          & CVPR‘21 & 12 & 24.2 & 50.3 & 20.3 & 8.8  & 20.4 & 30.2 & 39.4 & $40 \mathrm{M}$ \\
ViTDet-ViT-B\cite{li2022exploring}           & ECCV‘22 & 50 & 28.1 & 55.7 & 24.8 & 10.2 & 23.9 & 35.2 & 45.4 & $47 \mathrm{M}$ \\
\midrule
\textbf{Deformable-DETR-based} &&&&&&&&&& \\
\midrule
Deformable-DETR-R50\cite{zhu2020deformable}  & ICLR'21 & 50 & 19.2 & 44.8 & 13.7 & 6.3  & 15.4 & 24.9 & 34.2 & $42 \mathrm{M}$ \\
Conditional-DETR-R50\cite{meng2021-CondDETR} & ICCV'21 & 50 & 25.7 & 52.8 & 15.0 & 7.9  & 20.3 & 28.0 & 36.5 & $44 \mathrm{M}$ \\
DAB-DETR-R50\cite{liu2022dabdetr}            & ICLR'22 & 50 & 27.2 & 55.1 & 20.6 & 10.3 & 22.5 & 31.9 & 37.2 & $47 \mathrm{M}$ \\
DINO-R50\cite{zhang2022dino}                 & ICLR'23 & 36 & 28.9 & 59.4 & 22.4 & 12.5 & 22.7 & 34.7 & 42.8 & $47 \mathrm{M}$ \\
Co-DINO-R50\cite{codetr2022}                 & ICCV‘23 & 36 & 29.7 & 61.3 & 23.9 & 13.6 & 25.3 & 36.1 & 45.1 & $55 \mathrm{M}$ \\
\midrule
Baseline-R50 \cite{Lv2023DETRsBY}& - & 72                                       & 29.3 & 60.2 & 25.2 & 13.2 & 26.9 & 35.4 & 44.6 & $42 \mathrm{M}$ \\
Ours-R50 & - & 72                                           & $\textbf{31.4}$ & $\textbf{63.7}$ & $\textbf{28.1}$ & $\textbf{15.6}$ & $\textbf{29.7}$ & $\textbf{38.4}$ & $\textbf{46.4}$ & $43 \mathrm{M}$ \\
\bottomrule
\end{tabularx}
}
\label{soda-d}
\end{table*}

\section{Experiment}

\subsection{Dataset \& Metrics}

We perform quantitative analysis on small object datasets in driving scenes and aerial photography scenes. SODA-D \cite{Cheng2022TowardsLS} and VisDrone \cite{du2019visdrone}. SODA-D contains 24,828 carefully selected high-quality images associated with 278,433 instances in nine categories and annotated with horizontal bounding boxes. These images exhibit significant diversity in time periods, geographical locations, weather conditions, camera perspectives, etc., making them a significant strength of this dataset. VisDrone is a data set dedicated to the detection of drone-captured images, mainly targeting small-sized objects. It contains various environmental settings (urban and rural), scenes with different population densities, and a variety of objects, including pedestrians, vehicles, and bicycles. The dataset contains over 2.6 million manually annotated bounding boxes for common target objects.

The average precision (AP) is used to verify the performance in our experiments. The average AP across multiple IoU thresholds is between 0.5 and 0.95 with an interval of 0.05. Furthermore, AP$_{50}$ and AP$_{75}$ are calculated under a single IoU threshold of 0.5 and 0.75, respectively. The definition of AP across scales differs in Visdrone and SODA-D. In Visdrone, the scale is divided into small (S), medium (M) and large (L), including objects in the range of $(0,32^2]$, $(32^2,96^2]$, respectively. is $(96^2,\infty]$. In SODA-D, the scales are divided into extremely small (ES), relatively small (RS), generally small (GS) and normal (N), including those located at $(0,144]$, $(144,400]$, $(400,1024]$ and $(1024,2000]$.

\subsection{Setup}

We adopt RT-DETR \cite{Lv2023DETRsBY} as our baseline method and leverage its codebase. RT-DETR exhibits superior accuracy and speed compared to traditional CNN-based methods. During training, we use the AdamW optimizer with a base learning rate of 0.0002, weight decay configured to 0.0001, and a linear warm-up step lasting 2000 iterations. The learning rate of the backbone network will be adjusted accordingly. Additionally, we also used an exponential moving average (EMA) with an EMA decay rate of 0.9999 for stability. Our basic data augmentation strategy includes stochastic operations such as color distortion, flipping, resizing, expansion, and cropping.

\subsection{Main Results}

In this study, we conduct a thorough comparison of state-of-the-art end-to-end object detection methods. To ensure a fair comparison, we equip all baseline models with ResNet-50 as the underlying backbone architecture.

Our evaluation results clearly position our proposed method as a superior performer on both datasets considered. Notably, our method shows superior performance on the VisDrone dataset, achieving a significant overall average precision (AP) score of 32.3\%. Impressively, this performance outperforms the baseline model by a significant 2.9\%.

Similarly, on the SODA-D dataset, our method once again demonstrates its strength, achieving state-of-the-art performance with an overall AP of 31.3\%. Compared with the baseline model, our method shows significant superiority, significantly outperforming the baseline model by 1.7\%.

These results highlight the effectiveness and dominance of our proposed method in advancing object detection, especially in scenarios involving small objects. The state-of-the-art performance achieved on two datasets validates the superiority of our approach.

\begin{table}[hbt]
\centering
\caption{Ablation study on three strategies including scale-aligned target (ST), sample reweighting (SR), and sample point refinement (SPR)}
\resizebox{0.42\textwidth}{!}{%
\begin{tabular}{ccc|cc|ccc}
\hline 
ST & SR & SPR &AP & $\mathrm{AP}_{50}$  & $\mathrm{AP}_S$ & $\mathrm{AP}_M$ & $\mathrm{AP}_L$ \\ 
\hline
$\checkmark$ & &  & 29.9 & 58.1  & 20.5 & 46.9 & 58.2 \\
 &$\checkmark$ &  & 30.5 & 58.7  & 21.1 & 48.2 & 58.6 \\
 & &$\checkmark$ & 31.2 & 61.0 & 21.5 & 48.3 & 58.5  \\
$\checkmark$ & $\checkmark$ & & 31.8 & 62.1 & 22.8 & 49.5 & 59.8  \\
$\checkmark$ &$\checkmark$ & $\checkmark$  &$\textbf{32.3}$ & $\textbf{66.4}$ & $\textbf{23.7}$ & $\textbf{50.0}$ & $\textbf{60.2}$ \\
\hline
\end{tabular}
}
\label{component}
\end{table}

\begin{table}[hbt]
\centering
\caption{Ablation study on the effectiveness of different choices of $\beta$}
\resizebox{0.3\textwidth}{!}{%
\begin{tabular}{c|cccc}
\hline
$\beta$ & 0.3 & 0.5 & 0.73 & 0.9 \\
\hline
AP & 30.4 & 31.4 & $\textbf{32.3}$ & 31.9 \\
\hline
\end{tabular}
}
\label{beta}
\end{table}

\begin{table}[hbt]
\centering
\caption{Ablation study on the effectiveness of different choices of $\theta$}
\resizebox{0.3\textwidth}{!}{%
\begin{tabular}{c|cccc}
\hline
$\theta$ & 1.5 & 3 & 6 & 9 \\
\hline
$\mathrm{AP}$ & 30.6 & 30.9 & $\textbf{32.3}$ & 31.7 \\
\hline
\end{tabular}
}
\label{theta}
\end{table}

\begin{table}[!ht]
    \centering
    \caption{Ablation study on scaling factor $\alpha$ and suppression degree $\gamma$ }
    \begin{tabularx}{0.43\textwidth}{c|*{6}{X}}
    \hline
    $\gamma$ & \multicolumn{3}{c|}{1.5} & \multicolumn{3}{c}{1.75} \\
    \hline
    $\alpha$ & 0.25 & 0.5 & 0.75 & 0.25 & 0.5 & 0.75\\
    \hline
    AP  & 32.1 & $\textbf{32.3}$ & 31.5 & 30.8 & 30.6 & 30.2 \\
    \hline
    \end{tabularx}
    \label{alphagamma}
    \vspace{-10pt}
\end{table}

\begin{table}[ht]
\centering
\caption{Ablation study on the effectiveness of different choices of expansion rate $\eta$}
\resizebox{0.52\textwidth}{!}{%
\begin{tabular}{c|cc|ccc}
\hline 
$\eta$ & $\mathrm{AP}$ & $\mathrm{AP}_{50}$  & $\mathrm{AP}_S$ & $\mathrm{AP}_M$ & $\mathrm{AP}_L$ \\ 
\hline
(1.5, 1.4, 1.3, 1.2, 1.1, 1.0) & 32.2 & 64.8 & 23.4 & 50.6 & 60.1 \\
(1.5, 1.3, 1.2, 1.1, 1.05, 1.0)  & $\textbf{32.3}$ & $\textbf{66.4}$ & $\textbf{23.7}$ & 50.4 & $\textbf{60.2}$ \\
(2.0, 1.8, 1.6, 1.4, 1.2, 1.0)  & 31.9 & 65.1 & 21.6 & 50.9 & 60.9 \\
(2.0, 1.6, 1.4, 1.2, 1.1, 1.0) & 32.0 & 66.0 & 22.3 & 49.7 & 59.6 \\
\hline
\end{tabular}
}
\label{eta}
\end{table}

\begin{table}[hbt]
\centering
\caption{Generalizability of proposed scale-aligned target (ST) and sample reweighting (SR) across different state-of-the-art label assignment methods.}
\resizebox{0.41\textwidth}{!}{%
\begin{tabular}{c|cc|ccc}
\hline 
Method & $\mathrm{AP}$& $\mathrm{AP}_{50}$  & $\mathrm{AP}_S$ & $\mathrm{AP}_M$ & $\mathrm{AP}_L$ \\ 
\hline
GFL\cite{li2020generalized} & 29.7 & 60.7 & 23.7 & 47.7 & 52.8 \\
+ST  & 30.8 & 62.1 & 24.1 & 48.6 & 53.2 \\
+SR  & 31.6 & 64.5 & 23.7 & 49.8 & 60.2 \\
\hline
TOOD\cite{feng2021tood} & 29.5 & 59.4 & 21.7 & 46.6 & 55.9 \\
+ST & 30.3 & 61.1 & 22.5 & 46.8 & 55.2 \\
+SR  & 31.5 & 64.2 & 23.1 & 48.0 & 57.4 \\
\hline
\end{tabular}
}
\label{generalizability}
\end{table}

\subsection{Ablation Study}

\noindent\textbf{Individual Component Contribution} We perform a series of ablations on key components, namely task-decoupled reweighting and object-oriented losses. The experimental results are listed in Table \ref{component}. All components significantly contribute to the overall performance, improving objects of different sizes, with the most significant impact observed in the case of small objects. \par

\noindent\textbf{Hyperparameters} The parameter $\beta$ plays a crucial role in balancing the contribution of IoU score and scale score in the confidence objective. In a series of experiments detailed in Table \ref{beta}, we observe that when $\beta$ varies from 0.3 to 0.9, the overall performance initially increases to 32.3\% and then drops to 31.9\%. Therefore, we choose $\beta = 0.73$ as the best choice for our method. This suggests that scale information plays a crucial role in small object detection.

Similarly, $\theta$ acts as a scalar controlling factor, from area ratio to scale fraction, affecting the shape of the projection. In the experiments summarized in Table \ref{theta}, we found that when $\theta$ is between 1.5 and 9, the overall performance shows a trend of first increasing and then decreasing. Therefore, we choose $\theta = 6$ as the best choice for our method. \par

We also conduct experiments on suppression factors $\gamma$ and $\alpha$ to determine the optimal balance between positive and negative samples shown in Table \ref{alphagamma}. The results indicate substantial variations with $\gamma$, and the combination of $\gamma=1.5$ and $\alpha=0.5$ yields the best performance.

\begin{figure}[tb] 
    \centering 
    \includegraphics[width=1\linewidth]{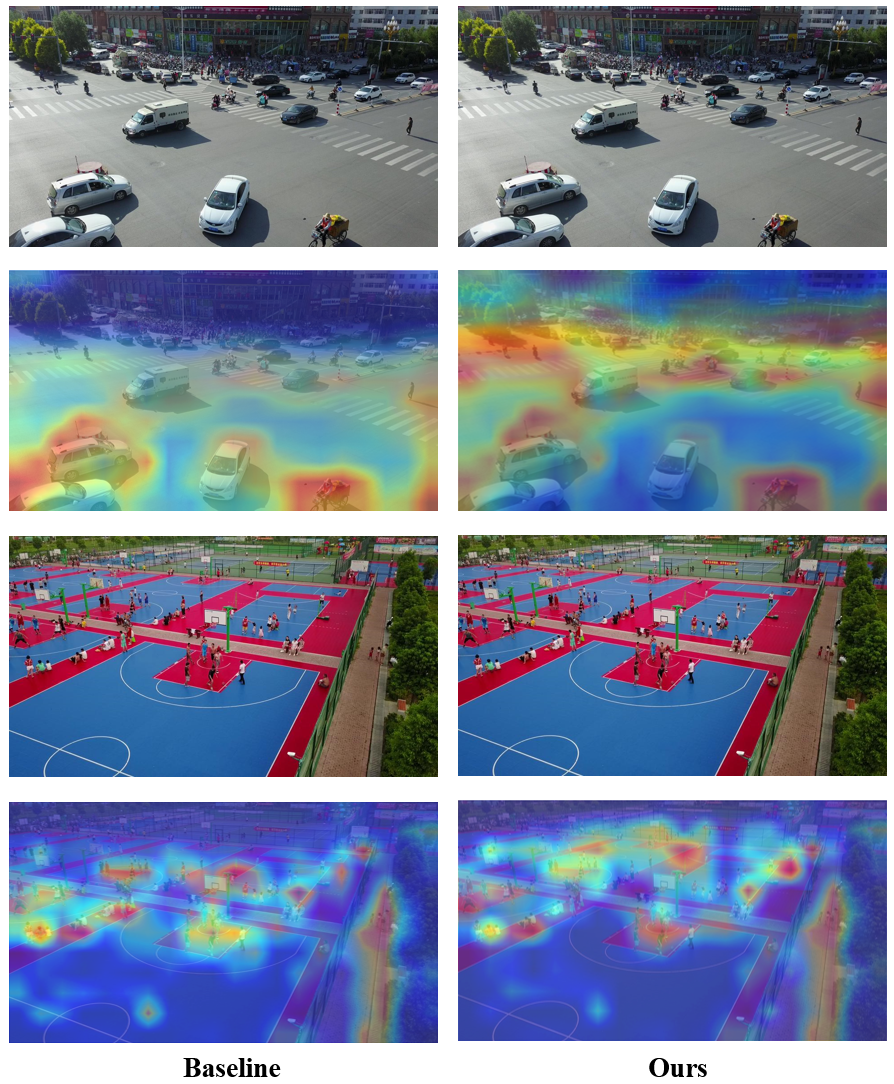} 
    \caption{Comparison of heatmap visualizations between the baseline and our method.}
    \label{cam}
\end{figure}
In our final experiment, we investigated the impact of various settings on the expansion rate $\eta$ in the six layers of the deformable attention module, as detailed in Table \ref{eta}. The settings involved six elements, each controlling the expansion rate. The first row represents a uniform reduction approach, while the second row represents an accelerated reduction method. The third and fourth rows involve a straightforward doubling of the expansion rate $\eta$. The setting of $(1.5, 1.3, 1.2, 1.1, 1.05, 1.0)$ shows the best performance of 32.3\%. One potential explanation for this behavior is that deformable attention may initially explore and identify the object in the early layers and subsequently refine the bounding box in the later layers.

\noindent\textbf{Generalization} To further verify the generalizability of our target design and target reweighting method, we apply our method to two other state-of-the-art label assignment methods GFL \cite{li2020generalized} and TOOD \cite{feng2021tood}. The experiments are based on our framework, as shown in Table \ref{generalizability}. The results demonstrate that our Scale-aligned Target proves to be a superior choice for small object detection, and the reweighting method facilitates the model in adaptively learning from data with class and size biases.

\subsection{Visualization}

Fig. \ref{cam} shows the visualization results of various methods applied to images in the VisDrone dataset. It is worth noting that our method shows quite superior performance compared with other methods. Initial numbers indicate that the baseline approach shows insufficient emphasis on extremely small targets. In contrast, our sampling method significantly enhances attention distribution, demonstrating enhanced accuracy and the ability to detect very small and faint targets effectively.

\section{Conclusion}

This paper presents sample point refinement, a novel approach to confine the localization and attention of sample points within deformable attention. This constraint ensures meaningful interactions within the region of interest while filtering out misleading information. Additionally, we introduce Scale-aligned Targets, incorporating scale information into target confidence, proving beneficial for small object detection. Furthermore, a task-decoupled sample reweighting mechanism guides the model to focus on challenging positive examples, adjusting classification loss based on difficulty assessed by a weight generator module. Extensive experiments on public datasets consistently demonstrate our approach outperforming state-of-the-art methods.

\newpage

\bibliographystyle{unsrt}
\bibliography{main}

\end{document}